\documentclass[letterpaper, 10 pt, conference]{ieeeconf}  %

\IEEEoverridecommandlockouts                              %

\usepackage{color,xcolor}
\usepackage{epsfig}
\usepackage{graphicx}

\usepackage{adjustbox}
\usepackage{array}
\usepackage{booktabs}
\usepackage{colortbl}
\usepackage{wrapfig}
\usepackage{hhline}
\usepackage{multirow}
\usepackage{subcaption} %
\usepackage[skip=1pt,labelfont=bf,font=footnotesize]{caption}
\usepackage{wrapfig}
\usepackage{floatflt}

\usepackage{amsmath,amsfonts,amssymb,amsthm}
\usepackage{mathtools}  %
\usepackage{bm}
\usepackage{nicefrac}
\usepackage{microtype}
\usepackage{times} %
\usepackage{mathptmx} %

\usepackage{changepage}
\usepackage{extramarks}
\usepackage{fancyhdr}
\usepackage{lastpage}
\usepackage{setspace}
\usepackage{soul}
\usepackage{xspace}
\usepackage{adjustbox}

\usepackage[breaklinks=true,colorlinks,citecolor=gray]{hyperref}

\usepackage{url}
\definecolor{urlblue}{rgb}{0,0,0.5}
\hypersetup{urlcolor=urlblue}
\usepackage[noadjust]{cite}

\usepackage{enumerate}
\let\labelindent\relax
\usepackage{enumitem}  %

\usepackage{titlesec}

\usepackage{makecell}

\usepackage{pifont} %

\usepackage[ruled,vlined]{algorithm2e}
\DontPrintSemicolon
\usepackage{framed}
\definecolor{shadecolor}{gray}{0.95}

\usepackage{xparse}
\usepackage{etoolbox}

\usepackage{courier}

\usepackage{amsmath,amsfonts,bm}

\def\eqref#1{equation~\ref{#1}}

\def\algref#1{algorithm~\ref{#1}}

\def\1{\bm{1}}

\DeclareMathAlphabet{\mathsfit}{\encodingdefault}{\sfdefault}{m}{sl}
\SetMathAlphabet{\mathsfit}{bold}{\encodingdefault}{\sfdefault}{bx}{n}

\newcommand{\stderr}[1]{\scriptsize $\pm #1$}

\newcolumntype{L}[1]{>{\raggedright\let\newline\\\arraybackslash\hspace{0pt}}m{#1}}
\newcolumntype{C}[1]{>{\centering\let\newline\\\arraybackslash\hspace{0pt}}m{#1}}
\newcolumntype{R}[1]{>{\raggedleft\let\newline\\\arraybackslash\hspace{0pt}}m{#1}}

\newcommand{\ours}{RAPID\xspace}

\newcommand{\fig}[1]{Fig.~\ref{#1}}

\newcommand{\ignore}[1]{}

\makeatletter
\DeclareRobustCommand\onedot{\futurelet\@let@token\@onedot}
\def\@onedot{\ifx\@let@token.\else.\null\fi\xspace}

\makeatother

\definecolor{MyDarkBlue}{rgb}{0,0.08,1}
\definecolor{MyDarkGreen}{rgb}{0.02,0.6,0.02}
\definecolor{MyDarkRed}{rgb}{0.8,0.02,0.02}
\definecolor{MyDarkOrange}{rgb}{0.40,0.2,0.02}
\definecolor{MyPurple}{RGB}{111,0,255}
\definecolor{MyRed}{rgb}{1.0,0.0,0.0}
\definecolor{MyGold}{rgb}{0.75,0.6,0.12}
\definecolor{MyDarkgray}{rgb}{0.66, 0.66, 0.66}
\definecolor{JiayuanColor}{rgb}{0.60,0.43,0.48}

\newif\ifpropositionfirstitem
\propositionfirstitemtrue

\newcommand{\xhdr}[1]{\noindent\textbf{#1}}
\newcommand{\Call}[2]{\text{\normalfont\textsc{#1}}\!\left(#2\right)}

\theoremstyle{plain}%

\title{\LARGE \bf
\ours: Robot Agentic Programming from Demonstrations 
}

\author{Yuyao Liu$^{1,2}$, Jiayuan Mao$^{3,\dagger}$, David Hsu$^{2,4,\dagger}$,
Leslie Pack Kaelbling$^{1,\dagger}$, and Tom\'as Lozano-P\'erez$^{1,\dagger}$%
\thanks{$^{\dagger}$ Equally advising.}%
\thanks{Corresponding author: Yuyao Liu (\texttt{yuyaoliu@mit.edu}).}
\thanks{Work done while Yuyao Liu was visiting NUS.}%
\\[0.5ex]
{\normalsize $^{1}$Massachusetts Institute of Technology \quad
$^{2}$National University of Singapore}\\
{\normalsize $^{3}$University of Pennsylvania \quad $^{4}$NVIDIA}
}

\begin{document}
\nocite{IEEEcontrol}

\maketitle
\thispagestyle{empty}
\pagestyle{empty}

\begin{abstract}

Coding agents have demonstrated enormous success in solving complex programming problems. To leverage their potential for robot systems, this work introduces \textbf{Robot Agentic Programming from Demonstrations (\ours)}, which automatically generates, verifies, and refines robot programs, given a \textrm{single visual human demonstration}. The iterative agentic loop of code refinement requires several key ingredients: (i) a testable task specification, (ii) action primitives for robot execution, and (iii) an interactive environment for program execution and verification. \ours infers all three from the demonstration automatically. To make the resulting program reusable beyond the demonstration setting, \ours uses an \textit{object-centric relational program representation} that focuses on the underlying structure of the demonstrated strategy rather than the specific motion \textit{per se}: it expresses the action primitives as trajectory-optimization programs that realize object-level motion effects, while composing them through relational constraints that capture scene-specific geometry at run time.  We evaluated \ours in simulation on eight challenging contact-rich nonprehensile manipulation tasks as well as general prehensile manipulation tasks in the LIBERO-Pro benchmark. We also successfully deployed it on a real Franka arm and evaluated on all eight nonprehensile tasks. In all experiments, \ours demonstrated strong performance, with generalization over object pose, shape, material, and environment. Website: \url{https://yuyaoliu.me/projects/rapid}.
\end{abstract}

\section{Introduction}
\begin{figure}[t]    
    \centering
    \includegraphics[width=\linewidth]{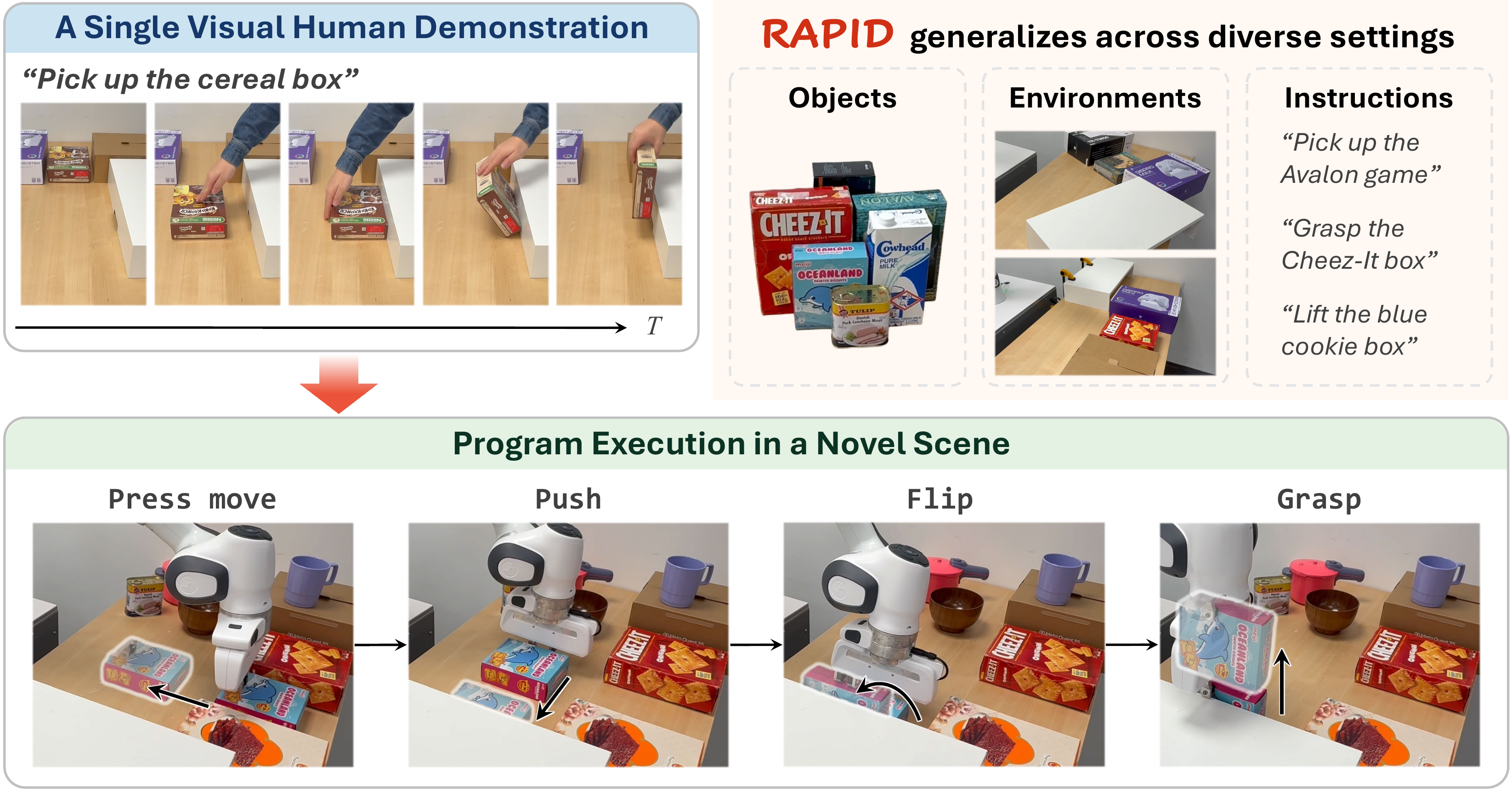}
    \caption{\textbf{\ours} (\textbf{R}obot \textbf{A}gentic \textbf{P}rogramm\textbf{i}ng from \textbf{D}emonstrations) generates a reusable manipulation program from a single visual human demonstration; the program generalizes across variations in object appearance, pose, geometry, physical properties, and scene configuration.}
    \vspace{-0.3in}
    \label{fig:teaser}
\end{figure}

Recently, coding agents~\cite{openai_codex,anthropic_claude_code} have shown remarkable capabilities in developing complex software systems, sparking growing interest in solving robotic manipulation tasks through programming~\cite{chen2026gap,xiao2026enpire,lu2026aspire,fu2026cap,berman2026claude,noematrix2026roborsi,liang2023code}. Consider how a human engineer would develop a robot program for the task in \fig{fig:teaser} before the advent of coding agents: given a \textit{task} (pick up the cereal box), they would implement a program \textit{representing} the solution strategy (press move$\to$push$\to$flip$\to$grasp), and \textit{verify} the program on test cases and iteratively debug it. This workflow mirrors the core strength of coding agents: not one-shot code generation, but the iterative cycle of generating, executing, verifying, and revising a program, commonly referred to as the agentic coding loop~\cite{ng2026three,huang2023agentcoder}.

However, bringing this agentic coding loop to robotic manipulation requires several ingredients that are not readily available: (a) a testable task specification that defines the intended outcome, (b) manipulation primitives that bridge program-level reasoning and low-level execution, and (c) interactive environments for verification. Existing robotic coding-agent systems typically assume these ingredients are provided \textit{a priori}, through ground-truth rewards and success signals, human-engineered primitives, and resettable environments. This raises a fundamental question: \textbf{how can we close the agentic coding loop without assuming that these ingredients are provided \textit{a priori}?}

\begin{figure*}[!t]
    \centering
    \includegraphics[width=\linewidth]{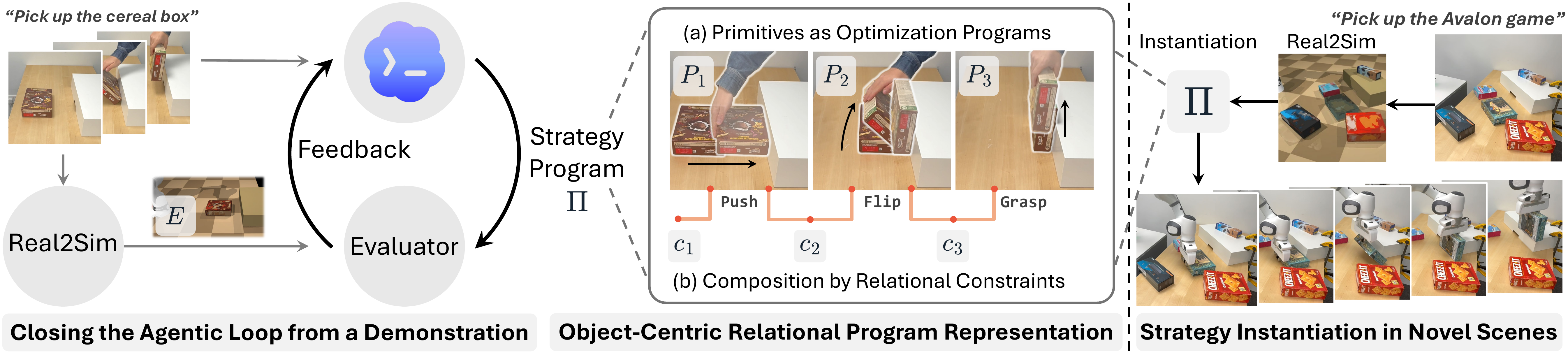}
    \caption{\textbf{Overview of \ours.} $E$ is a simulation environment reconstructed from the demonstration, $P_i$'s are manipulation primitives, $c_i$'s are relational constraints connecting two consecutive primitives, and $\Pi$ is the strategy program generated by the coding agent. }
    \label{fig:method_overview}
    \vspace{-0.3in}
\end{figure*}

Our key idea is to use \textit{demonstrations} as the programming interface for supplying the task- and motion-level information needed by the agentic coding loop. A demonstration directly shows a feasible way to accomplish a task and, together with a language description, conveys the intended outcome. Motivated by this perspective, we introduce \textbf{R}obot \textbf{A}gentic \textbf{P}rogramm\textbf{i}ng from \textbf{D}emonstrations (\textbf{\ours}), a framework for generating and iteratively refining robot programs from \textit{a single visual human demonstration} with a coding agent (\fig{fig:method_overview}). From the demonstration, \ours infers the task specification, constructs manipulation primitives and composes them into a strategy. It further reconstructs an interactive simulation environment from the demonstration for program verification and refinement, and generates scene variants to improve generalization. In this way, a single demonstration bootstraps the task specification, manipulation primitives, and verification environment for closing the agentic coding loop.

Rather than merely reproducing the demonstration, we aim to extract a reusable and generalizable manipulation program. Programs that directly encode poses, distances, or waypoints may succeed in the demonstrated scene, but can fail when the scene changes. In this work, we propose an \textit{object-centric relational program representation}, where the solution strategy is represented by object-centric relations and effects. Specifically, \ours writes primitives as local trajectory-optimization programs that realize object-level motion effects, such as pushing an object by a displacement or flipping it by an angle, while resolving scene-dependent geometry at execution time. At the task level, primitives are composed into a strategy through relational constraints that determine how the outcome of one phase specifies the next.  For example, in \fig{fig:program}, rather than replaying a fixed motion, the program first computes the goal position from the relation between the brown obstacle and the white fixture, and then determines how far to move the target object during the current primitive (press move) based on the state produced by the preceding primitive (flip). By expressing primitives and their composition relationally, \ours preserves the invariant structure of the demonstrated strategy while adapting to changes in object appearance, pose, geometry, physical properties, and scene configuration.

The remaining challenge is \textit{verification}: the coding agent needs an interactive environment in which candidate programs can be executed, evaluated, and revised. \ours reconstructs such an environment in simulation from the demonstration and augments it with scene variants generated by perturbing the demonstrated scene. The coding agent then generates task-specific filters that discard variants that are physically infeasible or inconsistent with the task. Afterwards, the coding agent executes candidate programs across this verification set and iteratively revises them based on observed failures. 

In this work, we focus on \textit{nonprehensile manipulation}, a particularly challenging domain in robotic manipulation. Many prehensile tasks can be composed from a compact and well-established set of reusable primitives~\cite{garrett2021integrated}, such as \textit{pick}, \textit{move in free space}, and \textit{place}. Nonprehensile manipulation, by contrast, involves a much broader range of geometry- and contact-dependent interactions, including pushing~\cite{mason2001mechanics}, pivoting~\cite{holladay2015general}, toppling~\cite{cheng2021contact}, flipping~\cite{odhner2013open}, and sliding~\cite{cheng2022contact}. Capturing this diversity with a fixed primitive library is difficult, making manual primitive design a substantial engineering burden. Nonprehensile manipulation therefore provides a natural testbed for studying whether a coding agent can construct reusable manipulation primitives from a single demonstration rather than relying on human engineering.

We evaluate \ours on eight contact-rich, multi-step nonprehensile manipulation tasks in simulation, asking whether programs constructed from a single demonstrated instance can generalize to substantial variations in object appearance, pose, geometry, physical properties, and scene configuration. \ours outperforms baselines without relying on pre-specified success signals, human-engineered manipulation primitives or provided resettable environments. To evaluate the generality of \ours beyond nonprehensile manipulation, we additionally test it on LIBERO-Pro~\cite{zhou2025libero}, a simulation benchmark featuring prehensile manipulation tasks. Finally, we demonstrate that \ours can be deployed effectively on a real robotic system.

\section{Related Work}

\xhdr{Programming for Robotics.}
Code as Policies~\cite{liang2023code} uses robot programs to connect language reasoning with control. Related work generates programs~\cite{singh2023progprompt,wang2025chainofmodality}, geometric constraints~\cite{huang2023voxposer,huang2024rekep}, rewards~\cite{ma2024eureka,xie2024text2reward}, and simulation environments~\cite{wang2024robogen}. Agentic programming adds iterative execution and refinement: CaP-X~\cite{fu2026cap} and related systems~\cite{xiao2026enpire,noematrix2026roborsi,chen2026gap,li2026roboclaw} improve programs through feedback, while Voyager~\cite{wang2023voyager} and skill-discovery approaches~\cite{lu2026aspire,zhang2026playful} grow reusable libraries. However, these approaches typically rely on task success signals, human-engineered primitives, or verification environments supplied \textit{a priori}. In contrast, \ours derives these ingredients from a single visual human demonstration to close the agentic coding loop.

\xhdr{Learning from Few Demonstrations.}
Learning from few demonstrations benefits from representations that preserve task-relevant structure across objects and scenes. Spatial abstractions support transfer through visual alignment~\cite{johns2021coarsetofine}, object geometry and correspondence~\cite{wen2022you,simeonov2022neural,shen2023distilled,zhu2024densematcher}, and contact or interaction structure~\cite{biza2023one,liu2025oneshot,sieb19graph,zhu2024orion}. Demonstrations can also be decomposed into subgoals and primitives~\cite{zhang2024universal,gao2024prime}. Structured skill representations enable long-horizon planning and composition through contact retargeting, symbolic models, or geometric reasoning~\cite{wu2024oneshot,liu2024learning,cheng2024nodtamp,nie2026learning,zandonati2025rational}. However, these representations are generally embedded in fixed learning or planning pipelines. In contrast, \ours combines an object-centric relational program representation of manipulation effects and composition constraints with agentic programming to iteratively construct, verify, and refine transferable manipulation strategies.

\xhdr{Nonprehensile Manipulation.}
 Nonprehensile manipulation exploits environmental contact to reposition objects and enable otherwise difficult grasps~\cite{chavandafle2014extrinsic}. Prior work develops mechanics-based models~\cite{lynch1996stable,holladay2015general}, contact-aware planners~\cite{lee2015hierarchical,cheng2021contact,cheng2022contact}, and optimization methods for trajectories and primitives~\cite{sleiman2019contact,aceituno2020global,pang2023global,huang2023autogenerated}. Learning-based approaches acquire contact-rich policies~\cite{zhou2023learning,zhou2023hacman,cho2024corn} or compose predefined primitives~\cite{nasiriany2022augmenting}. However, these methods typically rely on explicit task specifications or task-specific policy training. In contrast, \ours approaches nonprehensile manipulation through agentic programming from a single visual human demonstration, automatically constructing and iteratively refining the required primitives and their composition.

\section{Overview of \ours}
\label{sec:overview}

\xhdr{Problem Statement.}
We consider programming a reusable strategy for quasi-static rigid-body manipulation from \textit{a single visual human demonstration} $D$ and a language description $l$. The program should solve a family of task instances $\mathcal{E}$, each comprising a scene-instruction pair $(E,l)$, with similar objectives but varying object appearance, geometry, pose, physical properties, and scene configuration. No manipulation primitives, explicit success or reward signals, or instantiated simulation environments are provided \textit{a priori}. Given only $(D,l)$, the agent aims to construct executable primitives $\mathcal{P}=\{P_1,\ldots,P_N\}$, which produce low-level control signals, and compose them into a strategy $\Pi$ that captures the demonstration's underlying structure and generalizes to unseen task instances $(E^\star,l^\star)\in\mathcal{E}$:
\[
(D,l)
\mapsto
(\mathcal{P},\Pi),
\quad
\Pi(E^\star, l^\star;\mathcal{P}) \text{ succeeds for } (E^\star, l^\star) \in \mathcal{E}.
\]

\xhdr{Program Representation.}
We introduce an \textit{object-centric relational program representation} that enables strategies constructed by the agent to generalize across different scenes. \ours represents each primitive as a geometry- and physics-aware local trajectory-optimization program that achieves an object-level effect. A strategy then composes these primitives through relational constraints that specify the desired effect of each primitive. We detail this representation in Sec.~\ref{sec:relational_programs}.

\xhdr{Strategy Construction.}
As shown in \fig{fig:method_overview}, \ours closes the agentic coding loop by deriving task specifications, manipulation primitives, and verification environments from $(D,l)$. It segments the demonstration into several phases, infers a subgoal and success predicate for each phase, and performs real-to-sim reconstruction to obtain interactive simulation environments. \ours further synthesizes feasible scene variants for additional verification. For each phase, the coding agent generates a primitive and iteratively executes, verifies, and revises it across the reconstructed environment and its variants. \ours then infers the overall task goal and success predicate, composes the primitives into a strategy, and refines the composition through the same loop. We describe this construction and verification process in Sec.~\ref{sec:agentic_programming}.

\xhdr{Strategy Deployment.}
At deployment, given an RGB-D observation of a novel scene and a language instruction, \ours reconstructs the scene in simulation, instantiates the strategy in the reconstructed scene, and executes the resulting trajectory on the robot. Lightweight semantic binding assigns scene objects to the semantic roles in the strategy, while the strategy and primitives themselves remain unchanged.

\section{Object-Centric Relational Programs}
\label{sec:relational_programs}

We now define our object-centric relational program representation: primitives (Sec.~\ref{sec:relational_primitives}), their composition into task-level strategies (Sec.~\ref{sec:relational_strategy}), and strategy instantiation in novel scenes (Sec.~\ref{sec:novel_instantiation}).

\subsection{Primitives as Local Trajectory-Optimization Programs}
\label{sec:relational_primitives}
A primitive is a reusable behavior for realizing an object-level motion effect, such as pushing an object by a displacement or flipping an object by an angle. We represent a primitive as a local relational trajectory-optimization program
\begin{equation*}
    P_i=(G_i,C_i,\Psi_i,\rho_i,J_i)\in\mathcal{P}.
\end{equation*}
Here, $G_i$ is a semantic, object-centric description of the intended effect; $C_i$ is the relational interface; $\Psi_i$ is the verifiable success condition; $\rho_i$ is the geometry resolver; and $J_i$ is the optimization cost. Together, these components specify how the primitive is instantiated, optimized, and verified in a specific scene. A primitive is therefore neither a fixed trajectory nor a black-box policy. Table~\ref{tab:flip_primitive} illustrates each component for flipping an object against a fixture.

\begin{table}[t]
    \caption{\textbf{Local trajectory-optimization program for the primitive \textit{flip}.}}
    \label{tab:flip_primitive}
    \centering
    \begin{adjustbox}{width=\columnwidth}
    \setlength{\tabcolsep}{3pt}
    \begin{tabular}{>{\raggedright\arraybackslash}m{0.35\columnwidth}
                        >{\raggedright\arraybackslash}m{1.15\columnwidth}}
        \toprule
        \textbf{Component} & \textbf{Content} \\
        \midrule
        Semantic goal $G_i$
            & Flip the target object against a fixture by a specific angle. \\
        \addlinespace[2pt]
        \rowcolor{black!5}
        Relational interface \newline $C_i=(\mathbf{r}_i,\Gamma_i,\mathbf{q}_i)$
            & Semantic roles $\mathbf{r}_i$: (target object, support plane, fixture, manipulator). \newline
            Relations $\Gamma_i$: the object rests on the support plane, and the fixture exposes a vertical face oriented toward the target object and reachable by the manipulator. \newline 
            Effect parameter $\mathbf{q}_i$: the requested flipping angle.\\
        \addlinespace[2pt]
        Success predicate $\Psi_i$
            & Check that the achieved rotation matches the flipping angle and that the
              target object remains supported, near the fixture, and settled. \\
        \addlinespace[2pt]
        \rowcolor{black!5}
        Geometry resolver $\rho_i$
            & Resolve the support plane, the vertical face of the fixture, the flipping axis, and the dimensions of the target object at each step. \\
        \addlinespace[2pt]
        Final cost $J_i^{\mathrm{f}}$
            & Penalize terminal rotation error relative to the flipping angle, the gap between the fixture and the target object, and terminal velocity. \\
        \addlinespace[2pt]
        \rowcolor{black!5}
        Progress cost $J_i^{\mathrm{p}}$
            & Encourage rotation and manipulator proximity while discouraging
              backtracking and irregular motion. \\
        \bottomrule
    \end{tabular}
    \end{adjustbox}
    \vspace{-0.2in}
\end{table}

\xhdr{Relational Interface.}
We write the primitive interface as
\begin{equation*}
    C_i=(\mathbf{r}_i,\Gamma_i,\mathbf{q}_i).
\end{equation*}
Here, $\mathbf{r}_i$ declares the semantic roles involved in the primitive; for example, for the primitive \textit{flip} in Table~\ref{tab:flip_primitive}, these roles are \textit{(target object, support plane, fixture, manipulator)}. $\Gamma_i$ specifies the relations among these roles; for \textit{flip}, it encodes that the target object rests on the support plane and that the fixture exposes a vertical face oriented toward the target object and reachable by the manipulator. Finally, $\mathbf{q}_i$ declares the effect parameter; for \textit{flip}, it is the requested flipping angle.

To apply a primitive in a scene, an invocation supplies runtime arguments $(b_i,g_i)$ satisfying
\begin{equation*}
    b_i\in\operatorname{Bind}_E(\mathbf{r}_i,\Gamma_i),
    \quad
    g_i\in\operatorname{Val}(\mathbf{q}_i),
\end{equation*}
where $\operatorname{Bind}_E(\mathbf{r}_i,\Gamma_i)$ denotes the type-compatible assignments of the objects in the scene $E$ to $\mathbf{r}_i$ that satisfy $\Gamma_i$, and $\operatorname{Val}(\mathbf{q}_i)$ denotes the values matching the types declared by $\mathbf{q}_i$. Scene-dependent geometry remains internal to the primitive and is resolved by $\rho_i$ at execution time.

\xhdr{Geometry Resolver.}
Let $\mathcal{X}(E)$ be the state space of environment $E$. For the state $x_t\in\mathcal{X}(E)$ at time step $t$, the geometry resolver $\rho_i$ computes the geometric features $\phi_{i,t}$: 
\begin{equation*}
    \phi_{i,t} = \rho_i(E,x_t,b_i), \quad t=0,\ldots,T.
\end{equation*}
For example, for the primitive \textit{flip} in Table~\ref{tab:flip_primitive}, the agent would write detectors for the support plane, the vertical face of the fixture, the flipping axis, etc. 

\xhdr{Success Predicate.}
For a terminal state $x_T\in\mathcal{X}(E)$, the predicate $\Psi_i(E,x_T;g_i,\phi_{i,T})\in\{0,1\}$ provides an executable version of $G_i$. It checks the achieved object relations and requested effect, yielding an unambiguous signal for verification. 

\xhdr{Optimization Cost.}
For a simulated trajectory $\tau=(x_0,\ldots,x_T)$ in environment $E$, where $x_t\in\mathcal{X}(E)$, the primitive cost is
\begin{equation*}
    \begin{aligned}
        J_i(E,\tau;g_i,\phi_{i,0:T})
        = J_i^{\mathrm{f}}(E,x_T;g_i,\phi_{i,T}) 
        + \lambda_i J_i^{\mathrm{p}}(E,\tau;g_i,\phi_{i,0:T}).
    \end{aligned}
\end{equation*}
Here, $J_i^{\mathrm{f}}$ is the final cost, which measures how far $x_T$ is from the effect encoded by $G_i$; $J_i^{\mathrm{p}}$ is the progress cost, which may depend on the full trajectory and guides the optimizer through contact-rich interaction heuristics; and $\lambda_i\geq 0$ is a constant. The coding agent proposes the heuristics by reasoning about the visual human demonstration.

During primitive construction, the agent also proposes an initialization heuristic based on the resolved geometry and requested effect. The heuristic provides a plausible interaction trajectory from which to begin the search. Let $z$ parameterize a candidate robot motion, let $\tau(z)=\operatorname{Rollout}_E(x_0,z)$ denote its simulated state trajectory from $x_0$ in environment $E$, and let $\phi_{i,0:T}(z)$ be the feature sequence resolved along that trajectory. The executable trajectory is obtained as
\begin{equation*}
    z_i^\star
    = \arg\min_z
    J_i\!\left(E,\tau(z);g_i,\phi_{i,0:T}(z)\right).
\end{equation*}
We solve this trajectory-optimization problem with the cross-entropy method (CEM)~\cite{deboer2005tutorial}, a gradient-free sampling-based optimizer that allows us to optimize robot motions without differentiating through the contact-rich simulation. The primitive thereby adapts its motion to the object pose, size, geometry, physical properties, and requested effect.

\subsection{Strategies as Primitive Composition by Constraints}
\label{sec:relational_strategy}

\begin{figure}[t]
    \centering
    \includegraphics[width=\linewidth]{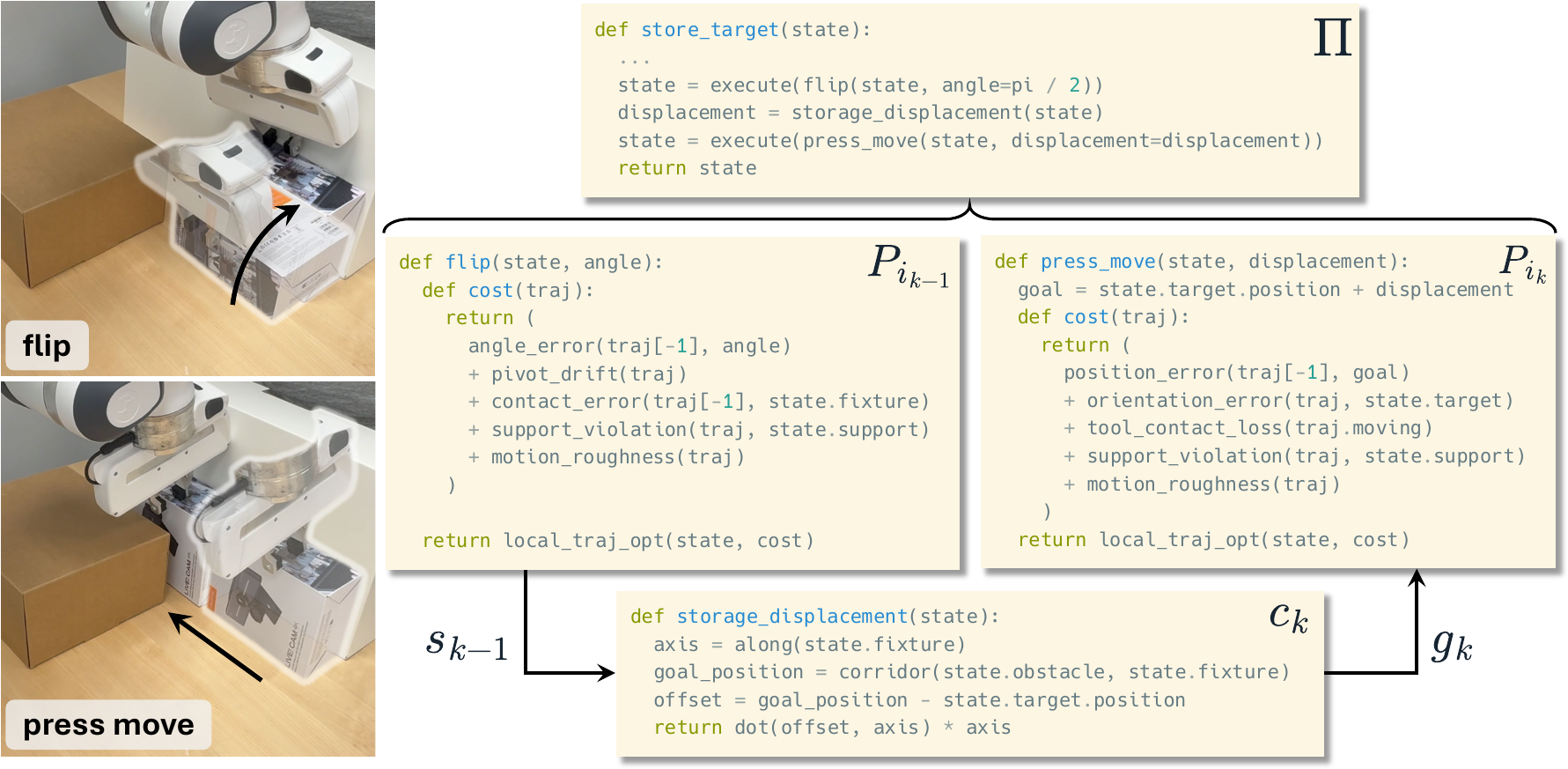}
    \caption{\textbf{Composing primitives by relational constraints.}}
    \label{fig:program}
    \vspace{-0.3in}
\end{figure}

After the primitives $\mathcal{P}$ are constructed, where each $P_i \in \mathcal{P}$ realizes an object-level motion effect, the next step is to compose them into a strategy $\Pi$, a reusable task-level program that accomplishes an overall task. It specifies the sequence of primitives to invoke, how task roles map to each primitive, and how the requested effect of each invocation is determined:
\begin{equation*}
    \Pi
    = \left(G_\Pi,C_\Pi,\Psi_\Pi,
      \left\langle(P_{i_k},m_k,c_k)\right\rangle_{k=1}^{K}\right),
\end{equation*}
where $G_\Pi$ describes the overall task goal and $\Psi_\Pi$ evaluates its final outcome.

\xhdr{Strategy Interface.}
A strategy exposes semantic roles and their required relations,
\begin{equation*}
    C_\Pi=(\mathbf{r}_\Pi,\Gamma_\Pi).
\end{equation*}
Invoking the strategy requires a semantic role binding $b\in\operatorname{Bind}_E(\mathbf{r}_\Pi,\Gamma_\Pi)$ that assigns an entity to each strategy role while satisfying $\Gamma_\Pi$. Primitive effect parameters are not exposed through $C_\Pi$; instead, they are derived inside the strategy during execution.

\xhdr{Primitive Composition by Constraints.}
A strategy consists of $K$ phases, each represented by a tuple $(P_{i_k},m_k,c_k)$, where $P_{i_k} \in \mathcal{P}$ is a primitive, $m_k$ is a role map, and $c_k$ is a connecting constraint. For the invoked primitive $P_{i_k}$, $m_k$ produces a role binding $b_k$ that fills the role declarations $\mathbf{r}_{i_k}$, while $c_k$ produces an effect argument $g_k$ that fills the parameter declarations $\mathbf{q}_{i_k}$. The connecting constraint declaratively specifies how phase $k$ depends on the outcome of phase $k-1$ and is evaluated just in time at the beginning of phase $k$:
\begin{equation*}
    \begin{aligned}
        b_k&=m_k(b)
            \in\operatorname{Bind}_E(\mathbf{r}_{i_k},\Gamma_{i_k}),\\
        g_k&=c_k(E,s_{k-1},b)
            \in\operatorname{Val}(\mathbf{q}_{i_k}),
    \end{aligned}
\end{equation*}
where $s_{k-1}\in\mathcal{X}(E)$ denotes the environment state after the execution of phase $k-1$. The phase is then executed as
\begin{equation*}
    s_k=\operatorname{Exec}\!\left(
        P_{i_k};E,s_{k-1},g_k,b_k\right).
\end{equation*}
Here, $\operatorname{Exec}$ solves the local trajectory-optimization problem of $P_{i_k}$ and executes the optimized trajectory.

For example, \fig{fig:program} shows a \textit{press move} following a \textit{flip}. The role map $m_k$ maps the strategy's target object, supporting surface, and tool roles to the corresponding roles of the \textit{press move} primitive, producing $b_k$. The connecting constraint $c_k$ determines the goal position from the relation between the brown obstacle and the white fixture. Using the target object's current position in the post-flip state $s_{k-1}$, it computes the displacement toward this goal along the fixture axis, producing the effect argument $g_k$ for the \textit{press move} primitive.

\subsection{Strategy Instantiation in Novel Scenes}
\label{sec:novel_instantiation}

To deploy a strategy in a novel scene, we only need to provide a semantic binding. Given a novel scene $E^\star$ and instruction $l^\star$, \ours uses a coding agent to rapidly generate a semantic binding $b^\star \in\operatorname{Bind}_{E^\star}(\mathbf{r}_\Pi,\Gamma_\Pi)$ that assigns scene entities to the semantic roles exposed by the strategy interface. 
Since this step only associates the exposed semantic roles with scene entities, it is lightweight and incurs little deployment overhead (a few seconds). At phase $k$, the strategy derives the primitive binding and effect argument automatically. 

Throughout this process, the strategy program $\Pi$ and the referenced primitives remain frozen. Generalization therefore arises from the object-centric relational program representation, together with the adaptation of trajectory optimization to the geometry and physical properties of the new scene.

\section{Agentic Programming from a Demonstration}
\label{sec:agentic_programming}

{
\makeatletter
{
  \let\oldalgopost\@algocf@post@ruled
  \def\@algocf@post@ruled{%
    \oldalgopost
    \vspace{-20pt}%
  }

\begin{algorithm}[t]
    \caption{\ours}
    \label{alg:agentic_programming}
    \small
    \renewcommand{\Call}[2]{\textnormal{#1}(#2)}
    \LinesNumbered
    \SetAlgoNoLine
    \SetAlgoNoEnd
    \SetCommentSty{textnormal}
    \SetKwComment{algref}{$\triangleright$ }{}
    \KwIn{Human demonstration $D$, instruction $l$}
    \KwOut{Verified primitives $\mathcal{P}$ and strategy $\Pi$}
    $\{(D_i,l_i)\}_{i=1}^{K}\gets\Call{Segment}{D,l}$\;
    $\mathcal{P}\gets\emptyset$\;
    \textcolor{black!70}{\textit{// Primitive construction}}\algref*[r]{Sec.~\ref{sec:primitive_construction}}
    \For{$i=1,\ldots,K$}{
        $\widehat{E}_i\gets\Call{Reconstruct}{D_i}$\algref*[r]{Sec.~\ref{sec:loop_ingredients}}
        $(G_i,\Psi_i)\gets\Call{InferObjective}{D_i,l_i}$\;
        $g_i\gets\Call{InferEffect}{D_i,G_i}$\;
        $P_i\gets\Call{ProposePrimitive}{D_i,G_i,\Psi_i,\widehat{E}_i}$\;
        $\mathcal{V}_i\gets\{(g_i,\widehat{E}_i)\}\cup\Call{Variants}{g_i,\widehat{E}_i,P_i}$\;
        $P_i\gets\Call{VerifyRevise}{P_i,\mathcal{V}_i,l_i}$\;
        $\mathcal{P}\gets\mathcal{P}\cup\{\Call{Freeze}{P_i}\}$\;
    }
    \textcolor{black!70}{\textit{// Strategy composition}}\algref*[r]{Sec.~\ref{sec:program_revision}}
    $\widehat{E}_0\gets\Call{Reconstruct}{D}$\;
    $\Pi\gets\Call{ComposeStrategy}{D,l,\mathcal{P},\widehat{E}_0}$\;
    $\mathcal{V}_\Pi\gets\{\widehat{E}_0\}\cup\Call{Variants}{\widehat{E}_0,\Pi}$\;
    $\Pi\gets\Call{VerifyRevise}{\Pi,\mathcal{V}_\Pi,l}$\;
    \KwRet{$\mathcal{P},\Call{Freeze}{\Pi}$}\;
    \textcolor{black!70}{\textit{// Shared verification and revision}}\;
    \SetKwProg{Fn}{function}{}{}
    \Fn{\textnormal{VerifyRevise}$(Q,\mathcal{V},a)$}{
        \Repeat{$\Call{AllSuccessful}{\mathcal{R}}$}{
            $\mathcal{R}\gets\Call{ExecuteEvaluate}{Q,\mathcal{V},a}$\;
            \If{$\neg\Call{AllSuccessful}{\mathcal{R}}$}{
                $Q\gets\Call{DiagnoseRevise}{Q,\mathcal{R}}$\;
            }
        }
        \KwRet{$Q$}\;
    }
\end{algorithm}
}
\makeatother
}

We now describe how \ours constructs and verifies the programs defined in Sec.~\ref{sec:relational_programs}, covering environment reconstruction, primitive construction, and strategy composition.

\subsection{Reconstructing Interactive Environments}
\label{sec:loop_ingredients}

\xhdr{Interactive Environment.}
To verify a manipulation program, \ours reconstructs an interactive simulation environment from the initial RGB-D frame of the demonstration. We first use a vision-language model (VLM) to identify the objects in the scene, assign semantic labels to them, and estimate their physical properties. Then, for each semantic name, we utilize SAM~3~\cite{carion2026sam} to get a segmentation mask. The RGB-D frame and the mask are then fed into SAM~3D~\cite{chen2026sam} to reconstruct a complete mesh. Finally, we use FoundationPose~\cite{wen2024foundationpose} to register each mesh to the RGB-D frame and assemble the reconstructed objects and robot model in MuJoCo~\cite{todorov2012mujoco}. The resulting environment preserves the scene entities, geometry, and spatial relations needed to execute the candidate program and evaluate its success predicates.

\xhdr{Feasible Scene Variants.}
To construct variants of the demonstrated scene, \ours uses a task-agnostic variation sampler to randomize object appearances, sizes, poses, and physical parameters such as mass, friction, and elasticity. However, the randomly sampled scenes might not be feasible for a certain primitive or strategy. Therefore, we ask the coding agent to write a filter for each primitive or strategy that accepts only variants preserving the declared roles and relations, task meaning, and physical feasibility.

\subsection{Primitive Construction}
\label{sec:primitive_construction}
Given the full demonstration $D$ and its natural language description $l$, \ours calls a subagent to partition $D$ into an ordered sequence of interaction segments $(D_1,\ldots,D_K)$ and assign a local description $l_i$ to each $D_i$. For each $(D_i,l_i)$, if the coding agent recognizes that the behavior is covered by an existing primitive in the primitive library $\mathcal{P}$, we can simply retrieve this primitive. Otherwise, the coding agent will infer a primitive specification consisting of a semantic goal $G_i$ and a success predicate $\Psi_i$ from $(D_i,l_i)$. It further reasons about the contact interactions that produce the demonstrated effect, the relational structure that should remain invariant across executions, and the quantities that might vary between invocations. For example, in Table~\ref{tab:flip_primitive}, the demonstrated rotation supplies the flipping-angle argument, while relations among the target object, support plane, fixture, and manipulator define the reusable interaction structure.

Given this specification and interaction analysis, the coding agent generates the relational interface $C_i$, geometry resolver $\rho_i$, and optimization cost $J_i$ introduced in Sec.~\ref{sec:relational_primitives}. As shown in Algorithm~\ref{alg:agentic_programming}, \ours verifies and revises $P_i$ across a set $\mathcal{V}_i$ containing the demonstrated effect--environment pair $(g_i,\widehat{E}_i)$ and feasible variants of both the environment and effect argument generated by a task-agnostic sampler. Verification continues until $\Psi_i$ is satisfied for every pair using its associated effect argument. The verified primitive is then added to the library $\mathcal{P}$.

\subsection{Strategy Composition}
\label{sec:program_revision}

After constructing the required primitives, \ours composes them into a task-level strategy. As shown in Algorithm~\ref{alg:agentic_programming}, given the full demonstration $D$ and instruction $l$, the coding agent infers a strategy specification consisting of the overall task goal $G_\Pi$ and final success predicate $\Psi_\Pi$. The sequence of the interaction segments $(D_1,\ldots,D_K)$ determines the ordering of the primitive composition. The agent then constructs the strategy interface $C_\Pi$ and, for each phase $k$, selects a verified primitive $P_{i_k}$, defines the role map $m_k$, and specifies the connecting constraint $c_k$. Together, these components form an initial version of the strategy $\Pi$. 

\ours verifies the strategy in the reconstructed environment $\widehat{E}_0$ and its feasible scene variants using the execution procedure in Sec.~\ref{sec:relational_strategy}. Failures guide the coding agent to revise the composition while keeping the verified primitives fixed, until $\Psi_\Pi$ is satisfied across this verification set $\mathcal{V}_\Pi$.

\section{Experiments}

In this section, we conduct experiments showing that (1) \ours successfully closes the agentic coding loop from a single visual human demonstration to construct reusable programs for nonprehensile manipulation (Table~\ref{tab:experiment_results}); (2) our object-centric relational program representation improves the generalization of the constructed manipulation programs (Tables~\ref{tab:experiment_results} and \ref{tab:libero_pro_results}); and (3) \ours can be effectively deployed on a real-world robotic system (\fig{fig:experiment_real}).

\begin{table*}[!t]
    \centering
    \caption{\textbf{Success rates on nonprehensile manipulation tasks in simulation.} We report mean$\pm$std across three trials; the best results are \textbf{bolded}.}
    \label{tab:experiment_results}
    {%
\small
\setlength{\tabcolsep}{7pt}
\renewcommand{\arraystretch}{1.08}
\begin{adjustbox}{max width=\linewidth}
\begin{tabular}{@{}l*{9}{c}@{}}
    \toprule
    \textbf{Method}
        & \textbf{Task 1}
        & \textbf{Task 2}
        & \textbf{Task 3}
        & \textbf{Task 4}
        & \textbf{Task 5}
        & \textbf{Task 6}
        & \textbf{Task 7}
        & \textbf{Task 8}
        & \textbf{Average} \\
    \midrule
    CaP
        & 0.053\stderr{0.009} & 0.000\stderr{0.000} & 0.000\stderr{0.000} & 0.000\stderr{0.000} & 0.000\stderr{0.000} & 0.000\stderr{0.000} & 0.013\stderr{0.009} & 0.040\stderr{0.016} & 0.013\stderr{0.003} \\
    CaP + OReP
        & 0.040\stderr{0.000} & 0.000\stderr{0.000} & 0.000\stderr{0.000} & 0.000\stderr{0.000} & 0.000\stderr{0.000} & 0.000\stderr{0.000} & 0.013\stderr{0.009} & 0.033\stderr{0.025} & 0.011\stderr{0.002} \\
    CaP-Agent0
        & 0.353\stderr{0.115} & 0.040\stderr{0.057} & 0.000\stderr{0.000} & 0.013\stderr{0.019} & 0.173\stderr{0.176} & 0.053\stderr{0.041} & 0.240\stderr{0.128} & 0.293\stderr{0.242} & 0.146\stderr{0.025} \\
    CaP-Agent0 + SV
        & 0.440\stderr{0.141} & 0.053\stderr{0.009} & 0.020\stderr{0.028} & 0.047\stderr{0.066} & 0.113\stderr{0.034} & 0.080\stderr{0.059} & 0.167\stderr{0.066} & 0.087\stderr{0.096} & 0.126\stderr{0.043} \\
    \ours w/o SV
        & \textbf{0.933\stderr{0.009}} & \textbf{0.833\stderr{0.125}} & 0.400\stderr{0.161} & 0.440\stderr{0.315} & 0.447\stderr{0.256} & 0.340\stderr{0.236} & 0.567\stderr{0.229} & 0.293\stderr{0.415} & 0.532\stderr{0.071} \\
    \textbf{\ours}
        & \textbf{0.933\stderr{0.057}} & \textbf{0.847\stderr{0.025}} & \textbf{0.700\stderr{0.033}} & \textbf{0.787\stderr{0.082}} & \textbf{0.793\stderr{0.009}} & \textbf{0.587\stderr{0.050}} & \textbf{0.807\stderr{0.203}} & \textbf{0.620\stderr{0.028}} & \textbf{0.759\stderr{0.024}} \\
    \bottomrule
\end{tabular}
\end{adjustbox}
}

    \vspace{-0.2in}
\end{table*}

\subsection{Experimental Setup for Nonprehensile Manipulation}
\xhdr{Tasks.} In this section, we evaluate \ours on eight nonprehensile manipulation tasks in simulation. For each task, we provide a single visual human demonstration together with a language description. In all tasks, the objects are initially not graspable by the parallel gripper, so the robot must execute a sequence of nonprehensile manipulation primitives, such as pushing, flipping, pivoting, and toppling, to reach the target configuration. Importantly, the primitives are \textit{not} provided to the coding agent \textit{a priori}. We provide the descriptions of the 8 tasks below, and \fig{fig:experiment_visualization} illustrates them in the real world:

\begin{figure}[t]
    \centering
    \includegraphics[width=\linewidth]{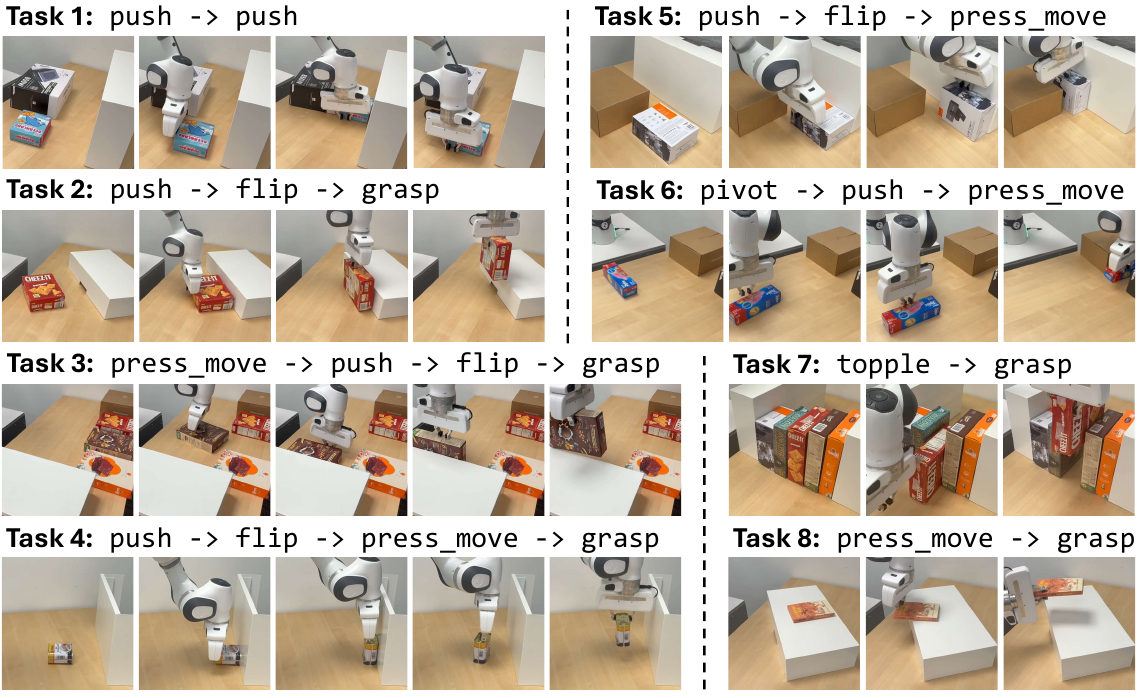}
    \caption{\textbf{Real-world execution of the solution trajectories found by \ours.}}
    \label{fig:experiment_visualization}
    \vspace{-0.3in}
\end{figure}

\xhdr{T1.} \textit{Push the object toward the corridor, switch the contact face, and push it again to store it inside the corridor.}

\xhdr{T2.} \textit{Push the object toward the fixture, flip it to expose a graspable edge, and grasp it.}

\xhdr{T3.} \textit{Move the object away from the obstacles, push it toward the fixture, flip it to expose a graspable edge, and grasp it.}

\xhdr{T4.} \textit{Push the object toward the fixture, flip it, reposition it to expose a graspable edge, and grasp it.}

\xhdr{T5.} \textit{Push the object toward the fixture, flip it to align with the opening, and move it into the opening for storage.}

\xhdr{T6.} \textit{Pivot the box to align it with the corridor, push it toward the corridor, and then move it to store it inside the corridor.}

\xhdr{T7.} \textit{Topple the object to expose an edge and grasp it.}

\xhdr{T8.} \textit{Move the object to the side of the platform to expose a graspable edge, and grasp it.}

\xhdr{Benchmark.} We construct a simulation benchmark in MuJoCo~\cite{todorov2012mujoco} using a Franka Research 3 robot arm equipped with a compliant operational-space controller~\cite{khatib1987unified}. Our primary evaluation metric is \textit{generalization}. For each of the 8 tasks, we construct 50 novel test scenes that differ from the demonstration scene in object pose, appearance, size, shape, and material. We also introduce irrelevant objects to make the test scenes visually distracting. We evaluate each method by measuring its success rate on each task using a private success criterion hidden from \ours. We run three trials of each experiment and report the mean and standard deviation.

\xhdr{Baselines.}
We compare six approaches along three principal axes: the closed agentic loop (\textbf{\textit{Agent}}), the \textbf{o}bject-centric \textbf{re}lational \textbf{p}rogram representation (\textbf{\textit{OReP}}), and synthesized \textbf{s}cene \textbf{v}ariants (\textbf{\textit{SV}}). 
\textit{1) Open-loop + Python (CaP)}: For each novel test scene, we use Code as Policies (CaP)~\cite{liang2023code} to generate a Python program conditioned on the demonstration, language description, and observation, without simulation feedback during program generation. 
\textit{2) Open-loop + OReP (CaP + OReP)}: We augment CaP with our OReP and cross-entropy method (CEM) optimizer, enabling motions to be expressed relative to scene entities. 
\textit{3) Agent + Python (CaP-Agent0)}: We use CaP-Agent0 from CaP-X~\cite{fu2026cap} to assess the benefit of our program representation. 
We give CaP-Agent0 a more favorable agentic-programming setting than that of \ours: during program construction, it receives rollout feedback consisting of ground-truth success signals, environment states, and visual observations; at evaluation time, it may additionally execute rollouts in each test scene and revise its program. 
\textit{4) Agent + Python + SV (CaP-Agent0 + SV)}: We further augment CaP-Agent0 with our scene-variant generation module, allowing it to refine its program on synthesized scene variants. 
\textit{5) Agent + OReP (\ours w/o SV)}: We remove the scene-variant generation module from \ours. 
\textit{6) Agent + OReP + SV (\ours)}: Our full method. 
Comparisons of 4) vs.\ 3) and 6) vs.\ 5) isolate the benefit of variant-based verification. We do not compare with ASPIRE~\cite{lu2026aspire} on nonprehensile tasks because it requires motion primitives to be provided. We use Codex~\cite{openai_codex} powered by GPT-5.6 Sol with high reasoning effort as the coding agent and provide the same robot control APIs for all methods.

\subsection{Simulation Experiments for Nonprehensile Manipulation}

\xhdr{\ours successfully constructs reusable programs by closing the agentic loop from a single demonstration.} As shown in Table~\ref{tab:experiment_results}, vanilla CaP performs poorly despite being allowed to generate a program tailored to each test scene. Without an interactive environment for verifying its programs, the agent cannot check whether the proposed motions produce the intended object interactions. Augmenting CaP with our object-centric relational program representation does not improve its performance either, suggesting that the representation alone is insufficient: without closed-loop verification, the coding agent struggles to reason about object relations and therefore fails to construct meaningful relational programs. In contrast, \ours reconstructs a simulation environment from the demonstration, infers a success predicate from the demonstration and language instruction, and uses this predicate to verify and iteratively refine candidate programs in simulation. In this way, \ours constructs a reusable program for each task that generalizes across diverse test scenes. In our experiments, this construction takes on the order of tens of minutes per task. This cost is incurred once per task: the resulting strategy and primitives remain fixed across the 50 novel scenes, with semantic role binding taking only a few seconds per scene.

\xhdr{The object-centric relational program representation enhances the generalization of the strategy program.} The comparison with the two CaP-Agent0 baselines in Table~\ref{tab:experiment_results} supports the benefit of our program representation: even with stronger verification access, including ground-truth feedback and evaluation-time refinement, CaP-Agent0 substantially underperforms \ours. Although CaP-Agent0 incorporates state feedback, its generated policies often use fixed world-frame offsets and predetermined timing, limiting spatial generalization and robustness to state deviations during execution. In contrast, \ours computes geometry-dependent subgoals from the state reached after each preceding phase. The gap is even larger for contact-rich, forceful behaviors such as flipping and toppling, which are naturally expressed by our relational, optimization-based motion primitives.

\xhdr{Ablation study of scene variant generation.} To evaluate the effectiveness of scene variant generation, we remove it from \ours and report the results in Table~\ref{tab:experiment_results}. Except for Tasks 1 and 2, the performance of \ours drops substantially, while the standard deviation of the success rate increases, indicating reduced stability and robustness. Interestingly, augmenting CaP-Agent0 with scene variant generation does not improve its performance. This suggests that, without our object-centric relational program representation, CaP-Agent0 struggles to construct a generalizable program that can solve diverse generated variants without overfitting to individual scenes.

\begin{table}[t]
    \centering
    \caption{\textbf{Experimental Results on LIBERO-Pro.}}
    \label{tab:libero_pro_results}
    {%
\small
\setlength{\tabcolsep}{6pt}
\renewcommand{\arraystretch}{1.08}
\begin{adjustbox}{max width=\linewidth}
\begin{tabular}{@{}l*{6}{c}@{}}
    \toprule
    \multirow{2}{*}{\textbf{Method}}
        & \multicolumn{2}{c}{\textbf{libero-object}}
        & \multicolumn{2}{c}{\textbf{libero-goal}}
        & \multicolumn{2}{c}{\textbf{libero-spatial}} \\
    \cmidrule(lr){2-3}\cmidrule(lr){4-5}\cmidrule(l){6-7}
        & \textbf{Pos. (Avg.)} & \textbf{Task. (Avg.)}
        & \textbf{Pos. (Avg.)} & \textbf{Task. (Avg.)}
        & \textbf{Pos. (Avg.)} & \textbf{Task. (Avg.)} \\
    \midrule
    OpenVLA     & 0            & 0             & 0             & 0             & 0             & 0             \\
    $\pi_0$     & 0            & 0             & 0             & 0             & 0             & 0             \\
    $\pi_{0.5}$ & 0.17         & 0.01          & 0.38          & 0             & 0.20          & 0.01          \\
    CaP-Agent0  & 0.22         & 0.18          & 0.26          & 0.17          & 0.12          & 0.14          \\
    ASPIRE      & \textbf{0.98}         & \textbf{0.95}          & \textbf{0.81} & 0.45          & 0.51          & 0.60          \\
    \ours       & \textbf{1}  & \textbf{0.95} & 0.67          & \textbf{0.53} & \textbf{0.83} & \textbf{0.81} \\
    \bottomrule
\end{tabular}
\end{adjustbox}
}

    \vspace{-0.25in}
\end{table}

\subsection{Simulation Experiments on LIBERO-Pro}
\xhdr{Experimental Setup.} To demonstrate the general applicability of \ours beyond nonprehensile manipulation, we evaluate it on LIBERO-Pro~\cite{zhou2025libero}, an extension of LIBERO~\cite{liu2023libero} with stronger position (Pos.) and instruction (Task.) perturbations. LIBERO-Pro predominantly features prehensile tasks, including pick-and-place and grasp-based articulated-object manipulation. We compare against two groups of baselines: (1) vision-language-action (VLA) models, including OpenVLA~\cite{kim2024openvla}, $\pi_0$~\cite{black2024pi_0}, and $\pi_{0.5}$~\cite{intelligence2025pi_}, whose results are directly taken from the LIBERO-Pro paper; and (2) code-as-policy (CaP) methods, including CaP-Agent0~\cite{fu2026cap} and ASPIRE~\cite{lu2026aspire}. For a fair comparison with CaP methods, which do not use demonstrations, \ours also receives no demonstrations and is given only the task description and a single debug scene. This setting is more restrictive than the setting used by ASPIRE, which uses 15 debug scenes per task. Moreover, while CaP baselines are provided with predefined motion primitives such as grasping, placing, and transporting, \ours must discover and implement these primitives solely through exploration and iterative refinement in the debug scene.

\xhdr{Results.} Table~\ref{tab:libero_pro_results} reports the results on LIBERO-Pro. \ours achieves the highest success rate on five of the six task settings, demonstrating the effectiveness of our object-centric relational program representation. On LIBERO-Goal (Pos.), \ours ranks second behind ASPIRE. This gap stems from articulated-object manipulation: primitives discovered by \ours from only a single debug scene, such as drawer opening, are less robust to spatial variation, whereas ASPIRE is provided with an \textit{interpolate segment} motion primitive that simplifies such interactions. Overall, these results demonstrate the generality of our framework beyond nonprehensile manipulation.

\subsection{Real-World Experiments for Nonprehensile Manipulation}
\begin{figure}[t]
    \centering
    \includegraphics[width=0.44\linewidth]{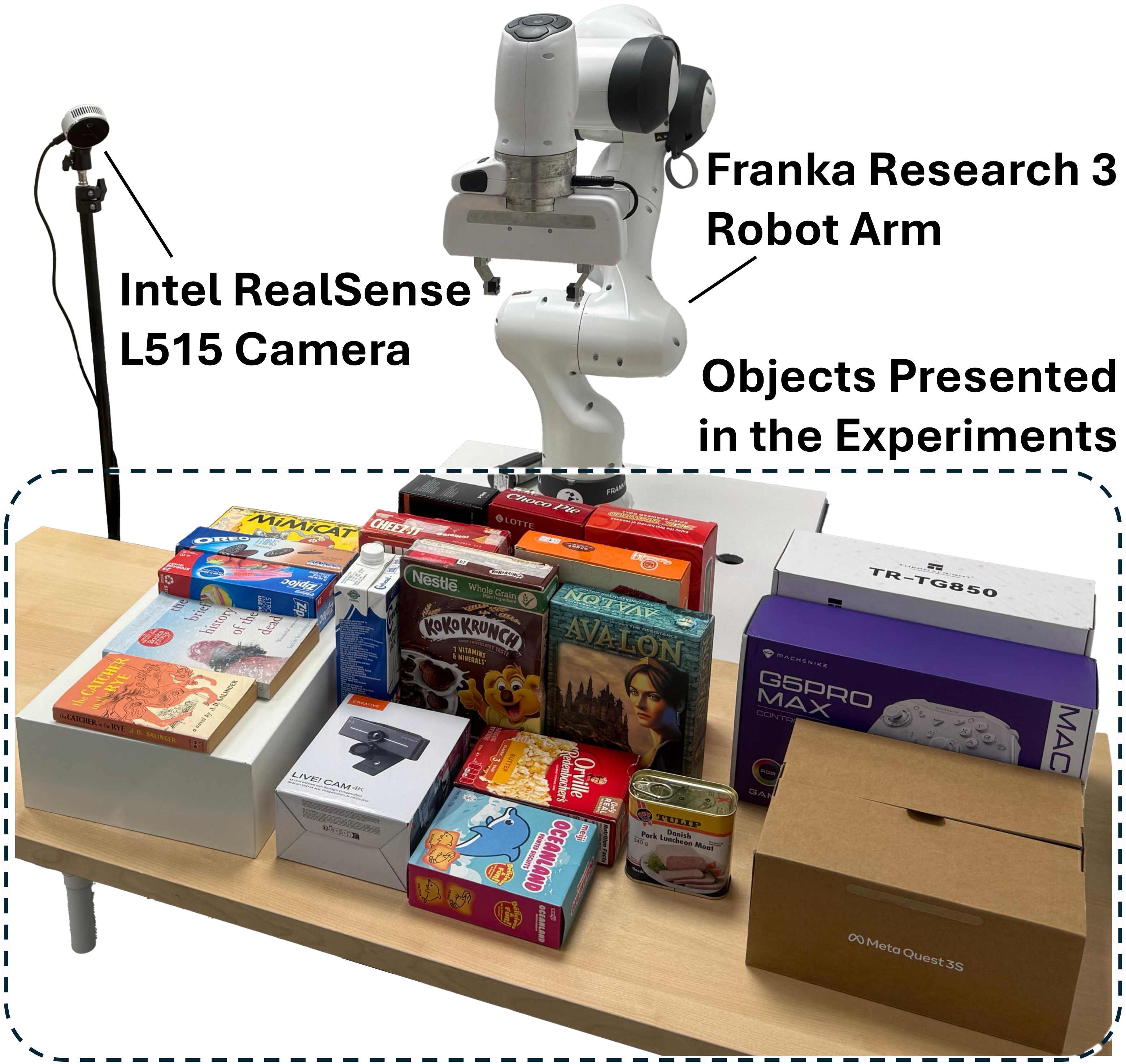}
    \includegraphics[width=0.54\linewidth]{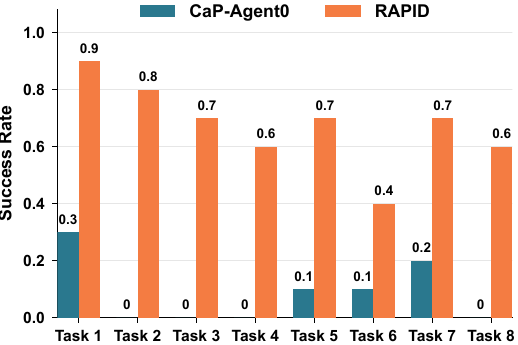}
    \caption{\textbf{Real-world experimental setup (left) and results (right).}}
    \label{fig:experiment_real}
    \vspace{-0.3in}
\end{figure}

To demonstrate that \ours can operate effectively on a real robotic system, we conduct real-world experiments on the eight nonprehensile manipulation tasks shown in \fig{fig:experiment_visualization}. As illustrated in \fig{fig:experiment_real}, our setup consists of a Franka Research 3 robot arm equipped with a parallel gripper and an Intel RealSense L515 camera for RGB-D observations. We compare \ours against CaP-Agent0~\cite{fu2026cap}. For each task, we evaluate 10 test scenes and report the success rate.

\xhdr{Results.} As shown in \fig{fig:experiment_real}, with our object-centric relational program representation and compliant operational-space controller, contact-rich nonprehensile behaviors optimized in simulation can be transferred directly to the real robot and executed effectively. In contrast, CaP-Agent0 struggles to generate reliable trajectories using procedural programs. Moreover, leveraging the VLM's prior knowledge to estimate physical parameters such as mass, friction, and restitution, together with SAM 3D's strong 3D reconstruction capability, helps keep the real-to-sim and sim-to-real gaps manageable.

\section{Conclusion}
We presented \ours, a framework that closes the agentic coding loop for robotic manipulation from a single visual human demonstration and a language description. By automatically deriving task specifications, manipulation primitives, and interactive verification environments, \ours enables a coding agent to construct, execute, and iteratively refine reusable manipulation strategies. Our object-centric relational program representation represents manipulation primitives as trajectory-optimization programs and composes them with relational constraints, enabling generalization to novel scenes. Experiments on eight contact-rich nonprehensile manipulation tasks demonstrate strong generalization across object and scene variations, while evaluations on LIBERO-Pro and a real robotic system support \ours's broader applicability and practical deployment.

In future work, \ours will incorporate other tools for learning primitives, such as reinforcement learning for manipulation with dexterous hands. Further, \ours will automatically benefit from the relentless progress in coding agents, robot foundation models, and simulation engines (e.g., for deformable objects~\cite{newton}), and scale up.

\bibliographystyle{IEEEtran}
\bibliography{references,references_new}

\begin{thebibliography}{10}
\providecommand{\url}[1]{#1}
\csname url@samestyle\endcsname
\providecommand{\newblock}{\relax}
\providecommand{\bibinfo}[2]{#2}
\providecommand{\BIBentrySTDinterwordspacing}{\spaceskip=0pt\relax}
\providecommand{\BIBentryALTinterwordstretchfactor}{4}
\providecommand{\BIBentryALTinterwordspacing}{\spaceskip=\fontdimen2\font plus
\BIBentryALTinterwordstretchfactor\fontdimen3\font minus
  \fontdimen4\font\relax}
\providecommand{\BIBforeignlanguage}[2]{{%
\expandafter\ifx\csname l@#1\endcsname\relax
\typeout{** WARNING: IEEEtran.bst: No hyphenation pattern has been}%
\typeout{** loaded for the language `#1'. Using the pattern for}%
\typeout{** the default language instead.}%
\else
\language=\csname l@#1\endcsname
\fi
#2}}
\providecommand{\BIBdecl}{\relax}
\BIBdecl

\bibitem{openai_codex}
{OpenAI}, ``{Codex},'' 2025, website: \url{https://openai.com/codex/}.

\bibitem{anthropic_claude_code}
{Anthropic}, ``{Claude Code},'' 2025, website:
  \url{https://claude.com/product/claude-code}.

\bibitem{chen2026gap}
K.~Chen \emph{et~al.}, ``{GaP: A Graph-as-Policy Multi-Agent Self-Learning
  Harness For Variational Automation Tasks},'' \emph{arXiv:2607.05369}, 2026.

\bibitem{xiao2026enpire}
W.~Xiao \emph{et~al.}, ``{ENPIRE: Agentic Robot Policy Self-Improvement in the
  Real World},'' \emph{arXiv:2606.19980}, 2026.

\bibitem{lu2026aspire}
R.~Lu \emph{et~al.}, ``{ASPIRE: Agentic /Skills Discovery for Robotics},''
  \emph{arXiv:2607.00272}, 2026.

\bibitem{fu2026cap}
L.~Fu \emph{et~al.}, ``{CaP-X: A framework for benchmarking and improving
  coding agents for robot manipulation},'' in \emph{ICML}, 2026.

\bibitem{berman2026claude}
S.~Berman, M.~Ilie, J.~Deng, and D.~Freeman, ``{Claude Plays Robotics},'' 2026,
  research Blog:
  \url{https://www.anthropic.com/research/claude-plays-robotics}.

\bibitem{noematrix2026roborsi}
{Noematrix Team}, ``{RoboRSI: Stable, Efficient, and Reusable Robot
  Self-Evolution in Complex Real-World Environments},'' 2026, research Blog:
  \url{https://lab.noematrix.ai/blog/2-roborsi/}.

\bibitem{liang2023code}
J.~Liang \emph{et~al.}, ``{Code as Policies: Language Model Programs for
  Embodied Control},'' in \emph{ICRA}, 2023.

\bibitem{ng2026three}
A.~Ng, ``{Three Key Loops for Building Great Software},'' 2026, research Blog:
  \url{https://www.deeplearning.ai/the-batch/three-key-loops-for-building-great-software}.

\bibitem{huang2023agentcoder}
D.~Huang \emph{et~al.}, ``{AgentCoder: Multi-Agent-Based Code Generation with
  Iterative\linebreak[4] Testing and Optimisation},'' \emph{arXiv:2312.13010},
  2023.

\bibitem{garrett2021integrated}
C.~R. Garrett \emph{et~al.}, ``{Integrated Task and Motion Planning},''
  \emph{Annual Review of Control, Robotics, and Autonomous Systems}, vol.~4,
  pp. 265--293, 2021.

\bibitem{mason2001mechanics}
M.~T. Mason, \emph{{Mechanics of Robotic Manipulation}}.\hskip 1em plus 0.5em
  minus 0.4em\relax MIT Press, 2001.

\bibitem{holladay2015general}
A.~Holladay, R.~Paolini, and M.~T. Mason, ``{A General Framework for Open-Loop
  Pivoting},'' in \emph{ICRA}, 2015.

\bibitem{cheng2021contact}
X.~Cheng, E.~Huang, Y.~Hou, and M.~T. Mason, ``{Contact Mode Guided
  Sampling-Based Planning for Quasistatic Dexterous Manipulation in 2D},'' in
  \emph{ICRA}, 2021.

\bibitem{odhner2013open}
L.~U. Odhner, R.~R. Ma, and A.~M. Dollar, ``{Open-Loop Precision Grasping with
  Underactuated Hands Inspired by a Human Manipulation Strategy},''
  \emph{T-ASE}, vol.~10, no.~3, pp. 625--633, 2013.

\bibitem{cheng2022contact}
X.~Cheng, E.~Huang, Y.~Hou, and M.~T. Mason, ``{Contact Mode Guided Motion
  Planning for Quasidynamic Dexterous Manipulation in 3D},'' in \emph{ICRA},
  2022.

\bibitem{zhou2025libero}
X.~Zhou \emph{et~al.}, ``{LIBERO-PRO: Towards Robust and Fair Evaluation of
  Vision-Language-Action Models beyond Memorization},''
  \emph{arXiv:2510.03827}, 2025.

\bibitem{singh2023progprompt}
I.~Singh \emph{et~al.}, ``{ProgPrompt: Generating Situated Robot Task Plans
  using Large Language Models},'' in \emph{ICRA}, 2023.

\bibitem{wang2025chainofmodality}
C.~Wang \emph{et~al.}, ``{Chain-of-Modality: Learning Manipulation Programs
  from Multimodal Human Videos with Vision-Language-Models},'' in \emph{ICRA},
  2025.

\bibitem{huang2023voxposer}
W.~Huang \emph{et~al.}, ``{VoxPoser: Composable 3D Value Maps for Robotic
  Manipulation with Language Models},'' in \emph{CoRL}, 2023.

\bibitem{huang2024rekep}
W.~Huang, C.~Wang, Y.~Li, R.~Zhang, and F.-F. Li, ``{ReKep: Spatio-Temporal
  Reasoning of Relational Keypoint Constraints for Robotic Manipulation},'' in
  \emph{CoRL}, 2024.

\bibitem{ma2024eureka}
Y.~J. Ma \emph{et~al.}, ``{Eureka: Human-Level Reward Design via Coding Large
  Language Models},'' in \emph{ICLR}, 2024.

\bibitem{xie2024text2reward}
T.~Xie \emph{et~al.}, ``{Text2Reward: Reward Shaping with Language Models for
  Reinforcement Learning},'' in \emph{ICLR}, 2024.

\bibitem{wang2024robogen}
Y.~Wang \emph{et~al.}, ``{RoboGen: Towards Unleashing Infinite Data for
  Automated Robot Learning via Generative Simulation},'' in \emph{ICML}, 2024.

\bibitem{li2026roboclaw}
R.~Li \emph{et~al.}, ``{RoboClaw: An Agentic Framework for Scalable
  Long-Horizon Robotic Tasks},'' \emph{arXiv:2603.11558}, 2026.

\bibitem{wang2023voyager}
G.~Wang \emph{et~al.}, ``{Voyager: An Open-Ended Embodied Agent with Large
  Language Models},'' \emph{TMLR}, 2024.

\bibitem{zhang2026playful}
J.~Zhang \emph{et~al.}, ``{Playful Agentic Robot Learning},''
  \emph{arXiv:2606.19419}, 2026.

\bibitem{johns2021coarsetofine}
E.~Johns, ``{Coarse-to-Fine Imitation Learning: Robot Manipulation from a
  Single Demonstration},'' in \emph{ICRA}, 2021.

\bibitem{wen2022you}
B.~Wen, W.~Lian, K.~Bekris, and S.~Schaal, ``{You Only Demonstrate Once:
  Category-Level Manipulation from Single Visual Demonstration},'' in
  \emph{RSS}, 2022.

\bibitem{simeonov2022neural}
A.~Simeonov \emph{et~al.}, ``{Neural Descriptor Fields: SE(3)-Equivariant
  Object Representations for Manipulation},'' in \emph{ICRA}, 2022.

\bibitem{shen2023distilled}
W.~Shen \emph{et~al.}, ``{Distilled Feature Fields Enable Few-Shot
  Language-Guided Manipulation},'' in \emph{CoRL}, 2023.

\bibitem{zhu2024densematcher}
J.~Zhu \emph{et~al.}, ``Densematcher: Learning 3d semantic correspondence for
  category-level manipulation from a single demo,'' in \emph{ICLR}, 2025.

\bibitem{biza2023one}
O.~Biza \emph{et~al.}, ``{One-shot Imitation Learning via Interaction
  Warping},'' in \emph{CoRL}, 2023.

\bibitem{liu2025oneshot}
Y.~Liu, J.~Mao, J.~B. Tenenbaum, T.~Lozano-P{\'e}rez, and L.~P. Kaelbling,
  ``{One-Shot Manipulation Strategy Learning by Making Contact Analogies},'' in
  \emph{ICRA}, 2025.

\bibitem{sieb19graph}
M.~Sieb, Z.~Xian, A.~Huang, O.~Kroemer, and K.~Fragkiadaki, ``{Graph-Structured
  Visual Imitation},'' in \emph{CoRL}, 2019.

\bibitem{zhu2024orion}
Y.~Zhu, A.~Lim, P.~Stone, and Y.~Zhu, ``{Vision-Based Manipulation from Single
  Human Video with Open-World Object Graphs},'' \emph{Autonomous Robots},
  vol.~50, no.~2, p.~27, 2026.

\bibitem{zhang2024universal}
Z.~Zhang \emph{et~al.}, ``{Universal Visual Decomposer: Long-Horizon
  Manipulation Made Easy},'' in \emph{ICRA}, 2024.

\bibitem{gao2024prime}
T.~Gao, S.~Nasiriany, H.~Liu, Q.~Yang, and Y.~Zhu, ``{PRIME: Scaffolding
  Manipulation Tasks with Behavior Primitives for Data-Efficient Imitation
  Learning},'' \emph{RA-L}, vol.~9, no.~10, pp. 8322--8329, 2024.

\bibitem{wu2024oneshot}
A.~Wu, R.~Wang, S.~Chen, C.~Eppner, and C.~K. Liu, ``One-shot transfer of
  long-horizon extrinsic manipulation through contact retargeting,'' in
  \emph{IROS}, 2024.

\bibitem{liu2024learning}
W.~Liu, N.~Nie, R.~Zhang, J.~Mao, and J.~Wu, ``{Learning Compositional
  Behaviors from Demonstration and Language},'' in \emph{CoRL}, 2024.

\bibitem{cheng2024nodtamp}
S.~Cheng, C.~R. Garrett, A.~Mandlekar, and D.~Xu, ``{NOD-TAMP: Generalizable
  Long-Horizon Planning with Neural Object Descriptors},'' in \emph{CoRL},
  2024.

\bibitem{nie2026learning}
N.~Nie \emph{et~al.}, ``{Learning Composable Skills by Discovering Spatial and
  Temporal Structure with Foundation Models},'' in \emph{ICRA}, 2026.

\bibitem{zandonati2025rational}
B.~Zandonati, T.~Lozano-P{\'e}rez, and L.~P. Kaelbling, ``{Rational Inverse
  Reasoning: Few-Shot Imitation by Inferring Intent through Planning},''
  \emph{arXiv:2508.08983}, 2025.

\bibitem{chavandafle2014extrinsic}
N.~Chavan-Dafle \emph{et~al.}, ``{Extrinsic Dexterity: In-Hand Manipulation
  with External Forces},'' in \emph{ICRA}, 2014.

\bibitem{lynch1996stable}
K.~M. Lynch and M.~T. Mason, ``{Stable Pushing: Mechanics, Controllability, and
  Planning},'' \emph{IJRR}, vol.~15, no.~6, pp. 533--556, 1996.

\bibitem{lee2015hierarchical}
G.~Lee, T.~Lozano-P{\'e}rez, and L.~P. Kaelbling, ``{Hierarchical Planning for
  Multi-Contact Non-Prehensile Manipulation},'' in \emph{IROS}, 2015.

\bibitem{sleiman2019contact}
J.-P. Sleiman, J.~Carius, R.~Grandia, M.~Wermelinger, and M.~Hutter,
  ``{Contact-Implicit Trajectory Optimization for Dynamic Object
  Manipulation},'' in \emph{IROS}, 2019.

\bibitem{aceituno2020global}
B.~Aceituno-Cabezas and A.~Rodriguez, ``{A Global Quasi-Dynamic Model for
  Contact-Trajectory Optimization in Manipulation},'' in \emph{RSS}, 2020.

\bibitem{pang2023global}
T.~Pang, H.~J.~T. Suh, L.~Yang, and R.~Tedrake, ``{Global Planning for
  Contact-Rich Manipulation via Local Smoothing of Quasi-Dynamic Contact
  Models},'' \emph{T-RO}, 2023.

\bibitem{huang2023autogenerated}
E.~Huang, X.~Cheng, Y.~Mao, A.~Gupta, and M.~T. Mason, ``{Autogenerated
  Manipulation Primitives},'' \emph{IJRR}, 2023.

\bibitem{zhou2023learning}
W.~Zhou and D.~Held, ``{Learning to Grasp the Ungraspable with Emergent
  Extrinsic Dexterity},'' in \emph{CoRL}, 2022.

\bibitem{zhou2023hacman}
W.~Zhou, B.~Jiang, F.~Yang, C.~Paxton, and D.~Held, ``{HACMan: Learning Hybrid
  Actor-Critic Maps for 6D Non-Prehensile Manipulation},'' in \emph{CoRL},
  2023, pp. 241--265.

\bibitem{cho2024corn}
Y.~Cho, J.~Han, Y.~Cho, and B.~Kim, ``{CORN: Contact-based Object
  Representation for Nonprehensile Manipulation of General Unseen Objects},''
  in \emph{ICLR}, 2024.

\bibitem{nasiriany2022augmenting}
S.~Nasiriany, H.~Liu, and Y.~Zhu, ``{Augmenting Reinforcement Learning with
  Behavior Primitives for Diverse Manipulation Tasks},'' in \emph{ICRA}, 2022.

\bibitem{deboer2005tutorial}
P.-T. de~Boer, D.~P. Kroese, S.~Mannor, and R.~Y. Rubinstein, ``{A Tutorial on
  the Cross-Entropy Method},'' \emph{Annals of Operations Research}, vol. 134,
  no.~1, pp. 19--67, 2005.

\bibitem{carion2026sam}
N.~Carion \emph{et~al.}, ``{SAM 3: Segment Anything with Concepts},'' in
  \emph{ICLR}, 2026.

\bibitem{chen2026sam}
X.~Chen \emph{et~al.}, ``{SAM 3D: 3Dfy Anything in Images},'' in \emph{CVPR},
  2026.

\bibitem{wen2024foundationpose}
B.~Wen, W.~Yang, J.~Kautz, and S.~Birchfield, ``{FoundationPose: Unified 6D
  Pose Estimation and Tracking of Novel Objects},'' in \emph{CVPR}, 2024.

\bibitem{todorov2012mujoco}
E.~Todorov, T.~Erez, and Y.~Tassa, ``{MuJoCo: A Physics Engine for Model-Based
  Control},'' in \emph{IROS}, 2012.

\bibitem{khatib1987unified}
O.~Khatib, ``{A Unified Approach for Motion and Force Control of Robot
  Manipulators: The Operational Space Formulation},'' \emph{IEEE Journal on
  Robotics and Automation}, vol.~3, no.~1, pp. 43--53, 1987.

\bibitem{liu2023libero}
B.~Liu \emph{et~al.}, ``{LIBERO: Benchmarking Knowledge Transfer for Lifelong
  Robot Learning},'' in \emph{NeurIPS}, 2023.

\bibitem{kim2024openvla}
M.~J. Kim \emph{et~al.}, ``{OpenVLA: An Open-Source Vision-Language-Action
  Model},'' in \emph{CoRL}, 2024.

\bibitem{black2024pi_0}
K.~Black \emph{et~al.}, ``{$\pi_0$: A Vision-Language-Action Flow Model for
  General Robot Control},'' in \emph{RSS}, 2025.

\bibitem{intelligence2025pi_}
{Physical Intelligence} \emph{et~al.}, ``{$\pi_{0.5}$: A Vision-Language-Action
  Model with Open-World Generalization},'' \emph{arXiv:2504.16054}, 2025.

\bibitem{newton}
{NVIDIA}, ``{Newton Physics Engine},'' 2025, website:
  \url{https://developer.nvidia.com/newton-physics}.

\end{thebibliography}

\end{document}